%% file: eacl2023.tex
\pdfoutput=1

\documentclass[11pt]{article}

\usepackage{EACL2023}

\usepackage{times}
\usepackage{latexsym}
\usepackage{graphicx}
\usepackage{booktabs}
\usepackage{multirow}
\usepackage{amsmath}
\usepackage{amsfonts}
\usepackage{subcaption}
\usepackage{tikz}
\usepackage{float}

\usepackage[T1]{fontenc}

\usepackage[utf8]{inputenc}

\usepackage{microtype}

\usepackage{inconsolata}

\title{Learning Disentangled Representations for Natural Language Definitions}

\input{sections/authors.tex}

\begin{document}
\maketitle

\input{sections/abstract.tex}
\input{sections/intro.tex}

\input{sections/method.tex}
\input{sections/eval.tex}

\input{sections/related.tex}

\input{sections/exp.tex}

\input{sections/conclusion.tex}

\input{sections_eacl_2023/limitations.tex}

\bibliography{references}
\bibliographystyle{acl_natbib}

\appendix

\input{sections/appendix.tex}

\end{document}

%% file: sections/authors.tex
\author{Danilo S. Carvalho$^1$ \quad Giangiacomo Mercatali$^1$$\dagger$ \quad Yingji Zhang$^1$$\dagger$ \quad Andre Freitas$^{1,2}$  \\
Department of Computer Science, University of Manchester, United Kingdom$^1$ \\
Idiap Research Institute, Switzerland$^2$ \\
\texttt{<firstname.lastname>@[postgrad.$\dagger$]manchester.ac.uk}}

%% file: sections/abstract.tex
\begin{abstract}
Disentangling the encodings of neural models is a fundamental aspect for improving interpretability, semantic control and downstream task performance in Natural Language Processing. Currently, most disentanglement methods are unsupervised or rely on synthetic datasets with known generative factors. We argue that recurrent syntactic and semantic regularities in textual data can be used to provide the models with both structural biases and generative factors. We leverage the semantic structures present in a representative and semantically dense category of sentence types, definitional sentences, for training a Variational Autoencoder to learn disentangled representations. Our experimental results show that the proposed model outperforms unsupervised baselines on several qualitative and quantitative benchmarks for disentanglement, and it also improves the results in the downstream task of definition modeling.


\end{abstract}



%% file: sections/intro.tex
\section{Introduction}

Learning disentangled representations is a fundamental step towards enhancing the interpretability of the encodings in deep generative models, as well as improving their downstream performance and generalization ability. Disentangled representations aim to encode the fundamental structure of the data in a more explicit manner, where independent latent variables are embedded for each generative factor~\citep{Bengio2013RepresentationLA}.

Previous work in machine learning proposed to learn disentangled representations by modifying the ELBO objective of the Variational Autoencoders (VAE)~\citep{Kingma2014AutoEncodingVB}, within an unsupervised framework~\citep{higgins2016beta,kim2018disentangling,chen2018isolating}. On the other hand, a more recent line of work claims the benefits of supervision in disentanglement~\citep{locatello2019challenging} and it advocates the importance of designing frameworks able to exploit structures in the data for introducing inductive biases. In parallel, disentanglement approaches for NLP have been tackling text style transfer, and evaluating the results with extrinsic metrics, such as style transfer accuracy~\citep{hu2017toward,john2019disentangled,cheng2020improving}.

\begin{figure}[t]
 \centering \includegraphics[scale=1]{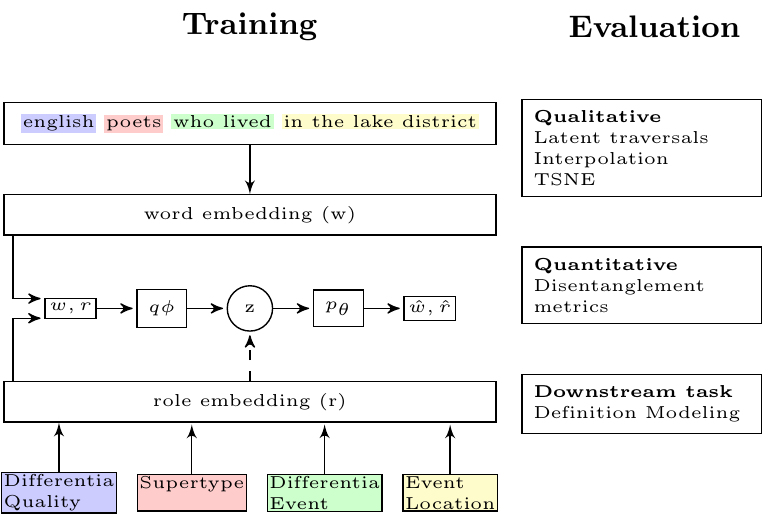} \caption{Left: Supervision mechanism with definition semantic roles (DSR) encoded in the latent space. The dotted arrow represent the conditional VAE version. Right: Evaluation framework.} \label{fig:diagram}
\end{figure}

While style transfer approaches investigate the ability to disentangle and control syntactic factors such as tense and gender, the aspect of understanding and disentangling the semantic structure in language is under-explored, but with recent attempts of separating syntactic and semantic latent spaces showing promising results~\citep{chen2019multi,bao2019generating}. Furthermore, evaluating disentanglement is challenging, because it requires knowledge of generative factors, leading most approaches to train on synthetic datasets~\citep{higgins2016beta,zhang2021unsupervised}.

In this work, we argue that recurrent semantic structures at sentence level can be leveraged both as inductive biases for enhancing disentanglement (\textbf{RQ1}) but also for providing meaningful generative factors that can be employed for evaluating the degree of disentanglement (\textbf{RQ2}). We also investigate whether organizing the generative factors in groups may facilitate learning and disentanglement (\textbf{RQ3}). As a result, this work focuses on natural language definitions, which are a textual resource characterised by a principled structure in terms of semantic roles, as demonstrated by previous work which proposed the extraction of structural and semantic patterns in this kind of data~\citep{silva2016categorization,silva2018recognizing}.

Seeking to address the highlighted issues and answer the research questions, we make the following contributions, also depicted in Figure~\ref{fig:diagram}.

1) We design a supervised framework for enhancing disentanglement in language representations by conditioning on the information provided by the semantic role labels (SRL) in natural language definitions. We present two mechanisms for injecting SRL biases into latent variables, firstly, reconstructing both words and corresponding SRL in a VAE, secondly, employing SRL information as input variables for a Conditional VAE~\citep{zhao2017learning}. 

2) We propose a framework for evaluating the disentanglement properties of the encodings on non-synthetic textual datasets. Our evaluation framework employs semantic role label groupings as generative factors, enabling the measurement of several contemporary quantitative metrics. The results show that the proposed bias injection mechanisms are able to increase the degree of disentanglement (separability) of the representations.

3) We demonstrate that models trained with our disentanglement framework are able to outperform contemporary baselines in the downstream task of definition modeling~\citep{noraset2017definition}.

%% file: sections/method.tex
\input{tikz/models.tex}

\section{Disentangling framework}
In this section we first describe the framework designed for improving disentanglement in natural language definitions with semantic role labels. Secondly, we present three models, shown in Figure~\ref{fig:vae_supervision} based on the Variational Autoencoder (VAE)~\citep{bowman2016generating} architecture for achieving disentanglement.

\subsection{Disentangling definitions}
\paragraph{Definition semantic roles}
Our framework is based on natural language definitions, which are a particular type of linguistic expression, characterised by high abstraction, and specific phrasal properties. Previous work in NLP for dictionary definitions~\citep{silva2018recognizing} has shown that there are categories that can be consistently found in most definitions. In fact, \citet{silva2018recognizing} define precise Semantic Role Labels (SRL) for phrases representing definitions, under the name of Definition Semantic Roles (DSR).

The example from~\citep{silva2018recognizing} classifies the semantic roles within "english poets who lived in the lake district" as follows. "poets" as noun category (supertype), "english" as quality of the term (Differentia Quality), "who lived" as event that the subject is involved with (differentia event), and "in the lake district" as the location of the action (Event location). The full DSRs proposed by~\citet{silva2018recognizing} are reported in Table~\ref{tab:srl_silva} in Appendix~\ref{sec:dsr_labels}.

\noindent
\textbf{Disentangling using SRL}~~~~Our goal is to enhance disentanglement in natural language by injecting categorical structures into latent variables. We find that this goal is well aligned with the findings of~\citet{locatello2019challenging}, where it is claimed that a higher degree of disentanglement may benefit from supervision and inductive biases. Our hypothesis is that we may leverage such semantic information for learning representation with higher degree of disentanglement. While in the context of this work we use dictionary definitions as a target empirical setting, we conjecture that these conclusions can be extended to broader definitional sentence-types. The core intuition behind the approach is that the supervision signal should increase the likelihood of point clustering in regions corresponding to, or related to the discrete supervision labels, given the network architecture formulation.

\subsection{Definition VAEs}

\textbf{Unsupervised VAE}~~~~The first training framework that we consider is the traditional variational autoencoder (VAE) for sentences~\citep{bowman2016generating}, which operates in an unsupervised fashion, as in Figure~\ref{fig:vae_architecture}. The unsupervised VAE employs a multivariate gaussian prior distribution $p(z)$ and generates a sentence $x$ with a decoder network $p_{\theta} (x|z)$. The joint distribution for the decoder is defined as $p(z) p_\theta (x|z)$, which, for a sequence of tokens $x$ of length $T$ result as $p_\theta (x|z) = \prod_{i=1}^T p_\theta (x_i | x_{<i}, z)$. The VAE objective consists into maximizing the expectation of the log-likelihood which is defined as $\mathbb{E}_{p(x)} \log p_\theta(x)$. Due to the computational intractability of the such expectation value, the variational distribution $q_\theta$ is employed to approximate $p_\theta (z|x)$.

As a result, an evidence lower bound $\mathcal{L}_\text{VAE}$ (ELBO) where $\mathbb{E}_{p(x)} [\log p_\theta(x)] \geq \mathcal{L}_\text{VAE}$, is derived as follows:
\[
\mathcal{L}_\text{Tokens} = 
\mathbb{E}_{q_\phi(z|x)} \Big[ \log p_{\theta} ( x | z ) \Big] - \text{KL} q_\phi(z|x) || p(z)
\]

\noindent
\textbf{DSR supervised VAE}~~~~The aim of this model is to inject the categorical structure of the definition semantic roles (DSR) into the latent variables, by factorizing them into the VAE auto-encoding objective function. In order to achieve this goal, we introduce the variable r for semantic roles, and train the "DSR VAE", where both sentence and semantic roles are auto-encoded. The variable $r$ here operates just as $x$, with the corresponding label values. As a result, two separate losses are produced and added together for the final loss, as shown in Figure~\ref{fig:dsr_vae_architecture}. The ELBO for semantic roles is defined as follows:
\[
\mathcal{L}_\text{Roles} =
\mathbb{E}_{q_\phi(z|r)} \Big[ \log p_{\theta} ( r | z ) \Big] - \text{KL} q_\phi(z|r) || p(z)
\]

The final loss is given by $\mathcal{L}_\text{Tokens} + \mathcal{L}_\text{Roles}$.\\

\noindent
\textbf{Conditional VAE with SRL}~~~~For explicitly leveraging the definition semantic roles, we propose a supervision mechanism based on the Conditional VAE (CVAE)~\citep{zhao2017learning}, shown in Figure~\ref{fig:cvae_architecture}. Similar to the previously described model, we instantiate a VAE framework, where $x$ is the variable for the tokens, and $r$ for the roles.
We perform auto-encoding for both roles and tokens, and additionally, we condition the decoder network on the roles. The CVAE is trained to maximize the conditional log likelihood of $x$ given $r$, which involves an intractable marginalization over the latent variable $z$.

The ELBO is defined as:
\begin{align*} \label{eq:dsr_cvae_loss}
\mathcal{L}_\text{CVAE} = &
\mathbb{E}_{q_\phi(z|r,x)} \Big[ \log p_{\theta} ( x | z,r ) \Big]  \\ \nonumber
& - \text{KL} q_\phi(z|x,r) || p(z|r)
\end{align*}

\paragraph{Training}
We consider LSTM-based VAE and Transformer-based VAE (Optimus ~\cite{li2020optimus}) as baselines. The training process follows the variational autoencoding methodology~\citep{Kingma2014AutoEncodingVB}. First, tokenization is performed in the sentences and the roles. The Encoder network involves feeding both first into embedding layers, then into LSTM / Transformer layers. Subsequently, two vectors $\mu$ and $\sigma$ are sampled with two linear layers, and the vector $z$ is computed with the re-parameterization trick. Finally, the decoder network is built with the LSTM / Transformer layers and another embedding layer, which return the same dimension that was given as input. 


%% file: tikz/models.tex

\begin{figure*}[h!]
\centering
\begin{subfigure}{0.3\textwidth}
    \includegraphics[width=\textwidth]{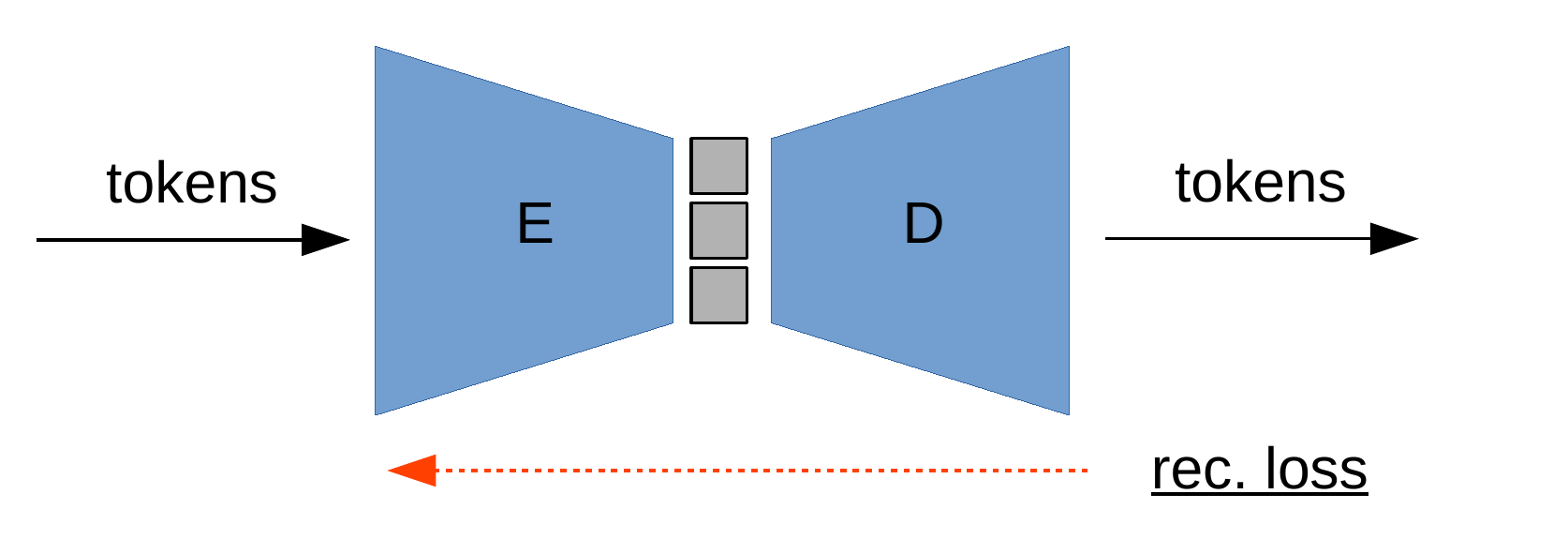}
    \caption{Unsupervised VAE}
    \label{fig:vae_architecture}
\end{subfigure}
\hfill
\begin{subfigure}{0.3\textwidth}
    \includegraphics[width=\textwidth]{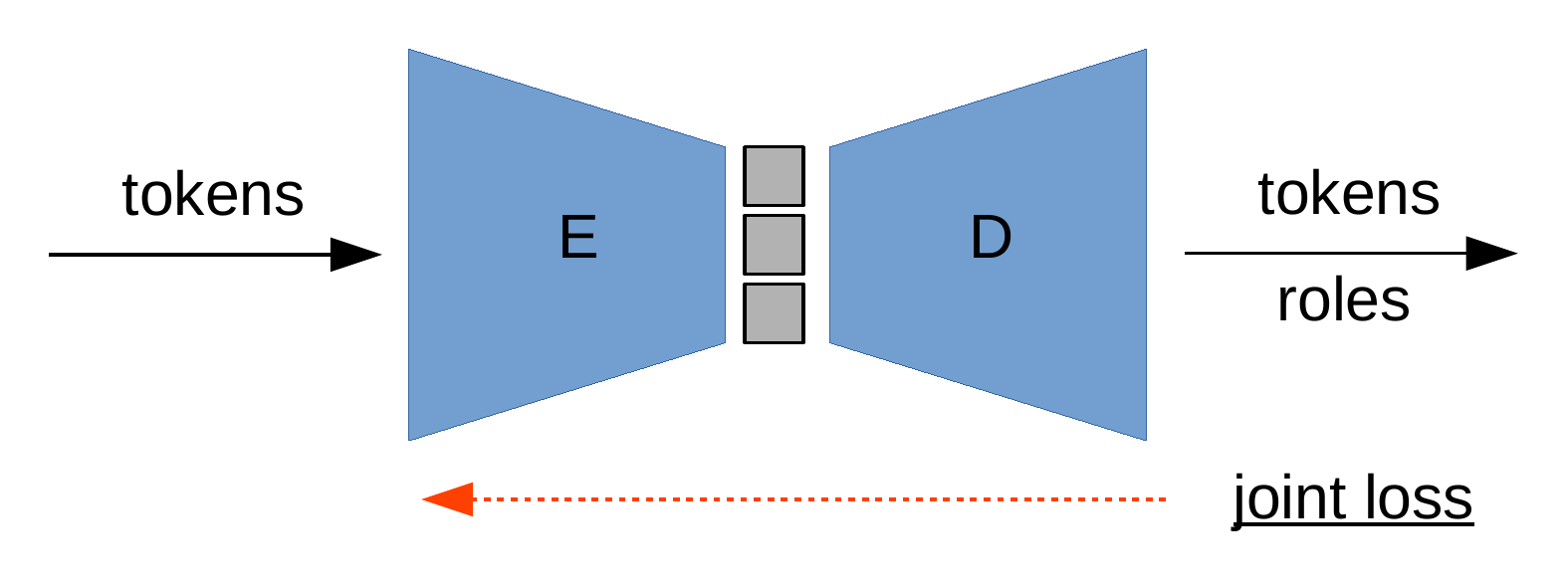}
    \caption{Supervised VAE}
    \label{fig:dsr_vae_architecture}
\end{subfigure}
\hfill
\begin{subfigure}{0.3\textwidth}
    \includegraphics[width=\textwidth]{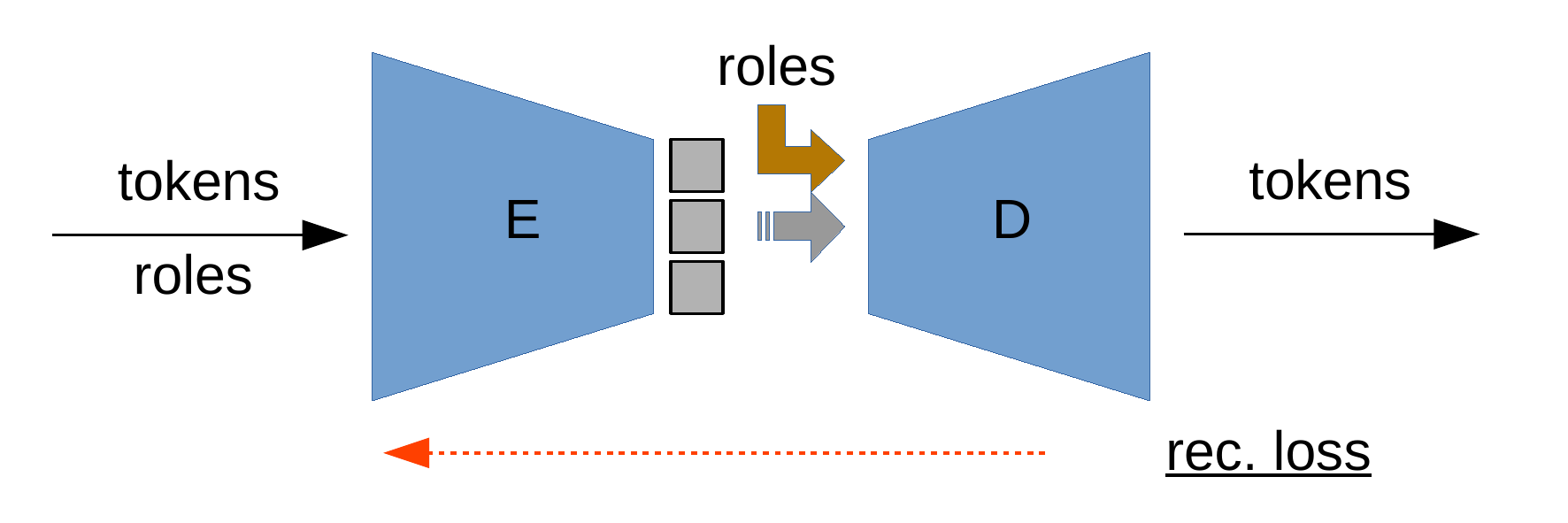}
    \caption{CVAE}
    \label{fig:cvae_architecture}
\end{subfigure}

\caption{Proposed architectures for learning disentangled representations in definitions.}
\label{fig:vae_supervision}
\end{figure*}


%% file: sections/eval.tex
\section{Evaluation framework} \label{sec:eval}
We first present the evaluation framework that for measuring disentanglement, then describe and justify the generative factor setup used in the experiments.

\vspace{-6pt}

\subsection{DSR as generative factors}
While early approaches for disentanglement in NLP have been proposed in the context of in style transfer applications~\citep{john2019disentangled,cheng2020improving} and are assessed purely in terms of style transfer accuracy, evaluating the intrinsic properties of the latent encodings is fundamental for disentanglement, as mentioned in several machine learning approaches~\citep{higgins2016beta,kim2018disentangling}. Recently, \citet{zhang2021unsupervised} proposed a framework for computing several popular quantitative disentanglement metrics such as~\citep{higgins2016beta,kim2018disentangling} testing it on synthetic datasets. The limitation in~\citep{zhang2021unsupervised} is that it works only with synthetic datasets.

\begin{figure}[ht]
    \centering
    \includegraphics[scale=.99]{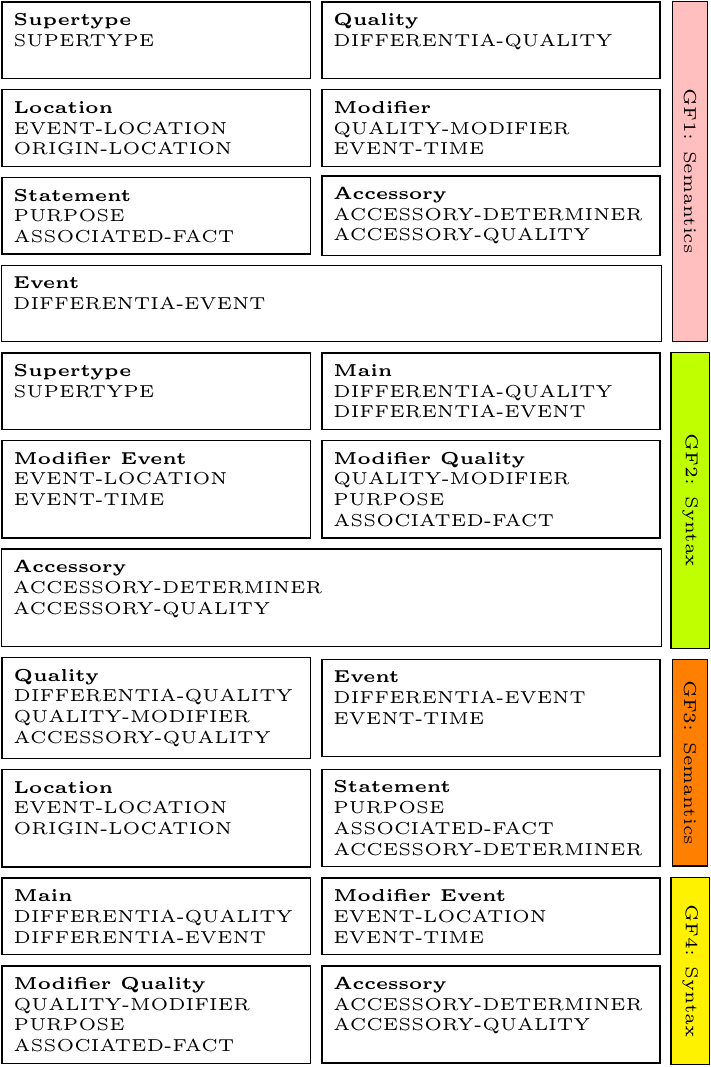}
    \caption{Generative factors for definitions.}
    \label{fig:generative_factors}
\end{figure}

In this work, we propose a method where semantic role labels, such as the ones provided in~\citep{silva2018recognizing}, are used as generative factors for evaluating the degree of disentanglement in the encodings. The framework, illustrated in Figure~\ref{fig:generative_factors}, considers multiple generative factors, where each factor is composed by a number of semantic roles (for example the factor "location" includes, origin-location, and event-location). In this way, the dataset can be seen as the result of a sampling of multiple generative factors, which is the same principle used when creating synthetic datasets for disentanglement. Once the generative factors are defined, the framework is enabled to compute a number of quantitative metrics for disentanglement, following the work from~\citet{zhang2021unsupervised}.

\subsection{Semantics and Syntax groups of DSR}
In order to categorize the definition semantic roles (DSR), we consider their structural and semantic dimensions in terms of their contribution to either the meaning (e.g., quality, location) or the structure (e.g., main terms, modifiers) of the definition sentence. We first create two DSR groups with semantic and two based on syntax, to evaluate which one would better facilitate disentanglement. For both syntax and semantic, we then create one group with "supertype" DSR and one without it, in order to understand the impact of the supertype DSR. The importance of "supertype" is due to its contribution to both abstraction groups and its predominant presence on the datasets analyzed ($\geq 97\%$).

\noindent
\textbf{Group 1: Semantics with Supertype}
Sets the factors in terms of their meaning, essentially abstracting categories of the DSRs, including the SUPERTYPE DSR as a single factor. Qualification, location, modification, declaration (statement) and supplementation (accessory) are semantic roles of a given term to its definition, which are described by the DSRs. 

\noindent
\textbf{Group 2: Syntax with Supertype}
Sets the factors in terms of their structural role in the definition sentence, including the SUPERTYPE DSR as a single factor. The ORIGIN-LOCATION DSR is omitted due to its syntactic overlap with EVENT-LOCATION and its low frequency in the datasets. 

\noindent
\textbf{Group 3: Semantics without Supertype}
Similar to group 1, but excluding the SUPERTYPE DSR, and repositioning the factor from \textit{modifier} and \textit{accessory} for higher abstraction. Relations of modification and supplementation (present in group 1) are suppressed to focus on lexical semantics, moving label ACCESSORY-DETERMINER to the declaratory group, EVENT-TIME to the event group and all quality related labels to the qualification group.

\noindent
\textbf{Group 4: Syntax without Supertype}
Similar to group 2, but excluding the SUPERTYPE DSR. Further abstractions are not conducted, as the definition roles already offer a stable structure for sentence construction.

%% file: sections/related.tex
\section{Related work}

\noindent
\textbf{Disentangled VAEs in language}
Early approaches in text disentanglement use VAEs with multiple adversarial losses for style transfer~\cite{hu2017toward,john2019disentangled}. More recently, \citet{cheng2020improving} propose a style transfer method which minimizing the mutual information between the latent and the observed variable, while \citet{colombo2021novel} propose an upper bound of mutual information for fair text classification. Disentanglement of syntactic and semantic information on sentences is explored by \citet{chen2019multi}, using multiple losses for word ordering and paraphrasing, and by \citet{bao2019generating} with linearized constituency tree losses. Finally, \citet{dupont2018learning} work on discrete factors for image models and the improvements in \citet{mercatali2021disentangling} proposed method for NLP lead to this work, where we move from the latter's implicit language features and LSTM-based architecture to explicit automatic annotations and a state-of-the-art Transformer-based architecture.
We focus our efforts into the representation of definitions, and propose to promote disentanglement by using biases provided as semantic roles, designing two VAE models to inject structural semantic information into the representation.
As an alternative architecture for generative modeling, Generative Adversarial Network (GAN) was not employed for this problem due to the non-contrastive nature of the input data (trying to leverage informed structural knowledge) and the emphasis on disentanglement as a mechanism to understand separability and control.


\noindent
\textbf{Disentanglement Evaluation}
\citet{vishnubhotla2021evaluation} evaluate disentanglement in synthetic text on various NLP tasks such as classification, retrieval and style transfer. \citet{zhang2021unsupervised} evaluate disentanglement of various VAE models on synthetic datasets where generative factors are known. Differently from these methods, we propose a new framework to evaluate non-synthetic natural language, where semantic role labels are used as generative factors. We model linguistic features of natural language definitions, with the goal of exploring the semantic properties that are encapsulated in it.


\noindent
\textbf{Definition models}
Early approaches in definition encoding include~\citep{hill2016learning}, which propose the first neural embedding model for dictionaries, and~\citep{bahdanau2017learning}, which present an RNN-based encoder decoder architecture for textual entailment and reading comprehension. More recently, methods based on Autoencoders~\citep{bosc2018auto} and transformers~\citep{Tsukagoshi2021DefSentSE} have been proposed. Various approaches for the task of generating a definition from a word (Definition Modeling) have been proposed, including RNN-based methods~\citep{noraset2017definition}, soft attention mechanisms~\citep{gadetsky2018conditional}, and span-based encoding schemes~\citep{bevilacqua2020generationary}. The semantic aspect of natural language definitions are explored in~\citep{silva2016categorization,silva2018recognizing}, where the concept of definition semantic roles is proposed.

%% file: sections/exp.tex
\section{Empirical analysis}
In this section, we firstly describe the empirical setup for experiments, secondly, we provide qualitative evaluation and thirdly, we measure various quantitative metrics. Finally, we demonstrate the capacity of the proposed models in the downstream task of definition modeling. 

\input{sections/exp_setup.tex}

\input{sections/qualitative.tex}

\begin{figure*}[ht!]
    \centering
    \begin{subfigure}{.5\textwidth}
        \hspace{-3cm}
        \includegraphics[scale=0.35]{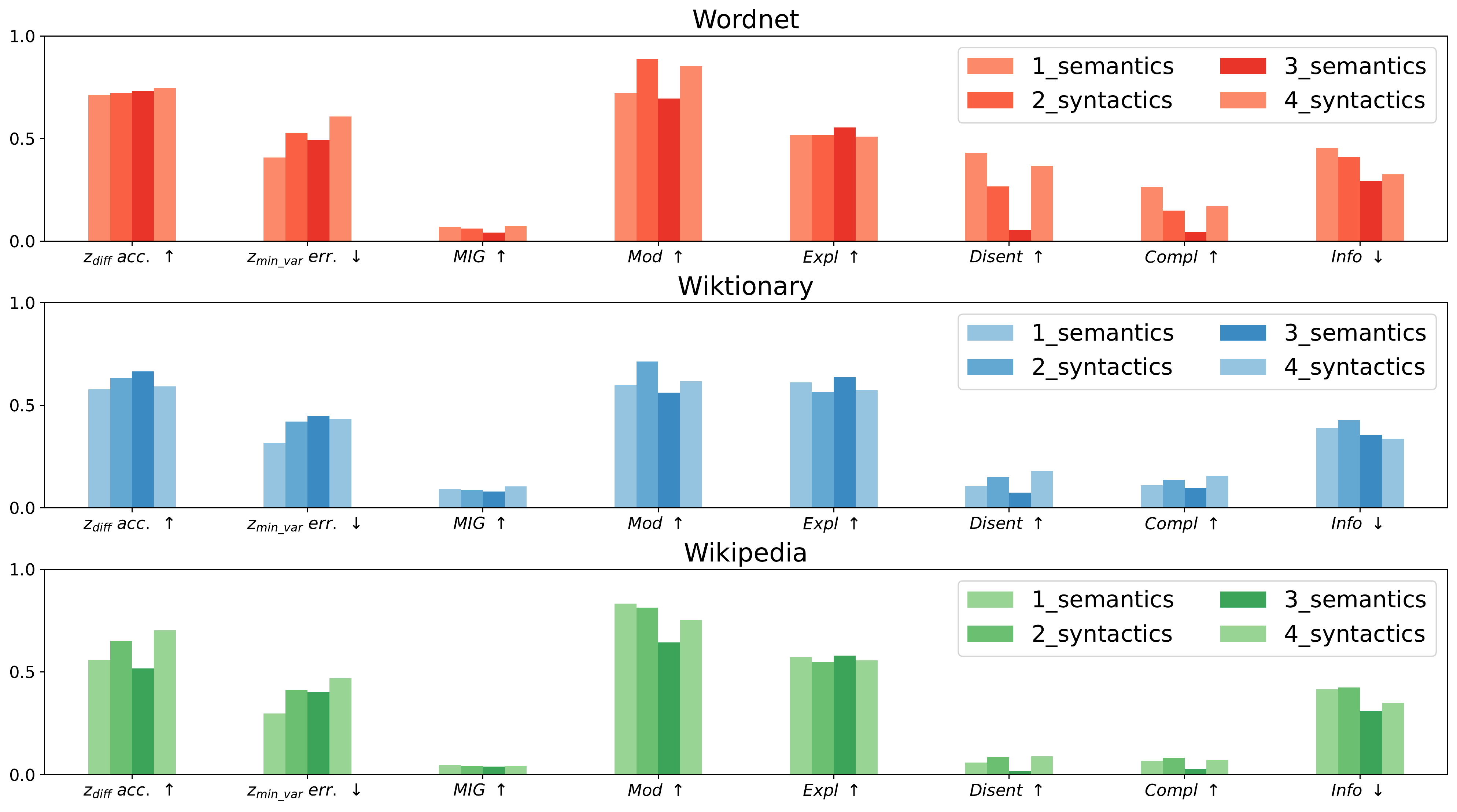}
        \caption{By generative factor}
        \label{fig:zdim_analysis}
    \end{subfigure}
    \begin{subfigure}{.5\textwidth}
        \hspace{-3cm}
        \includegraphics[scale=0.35]{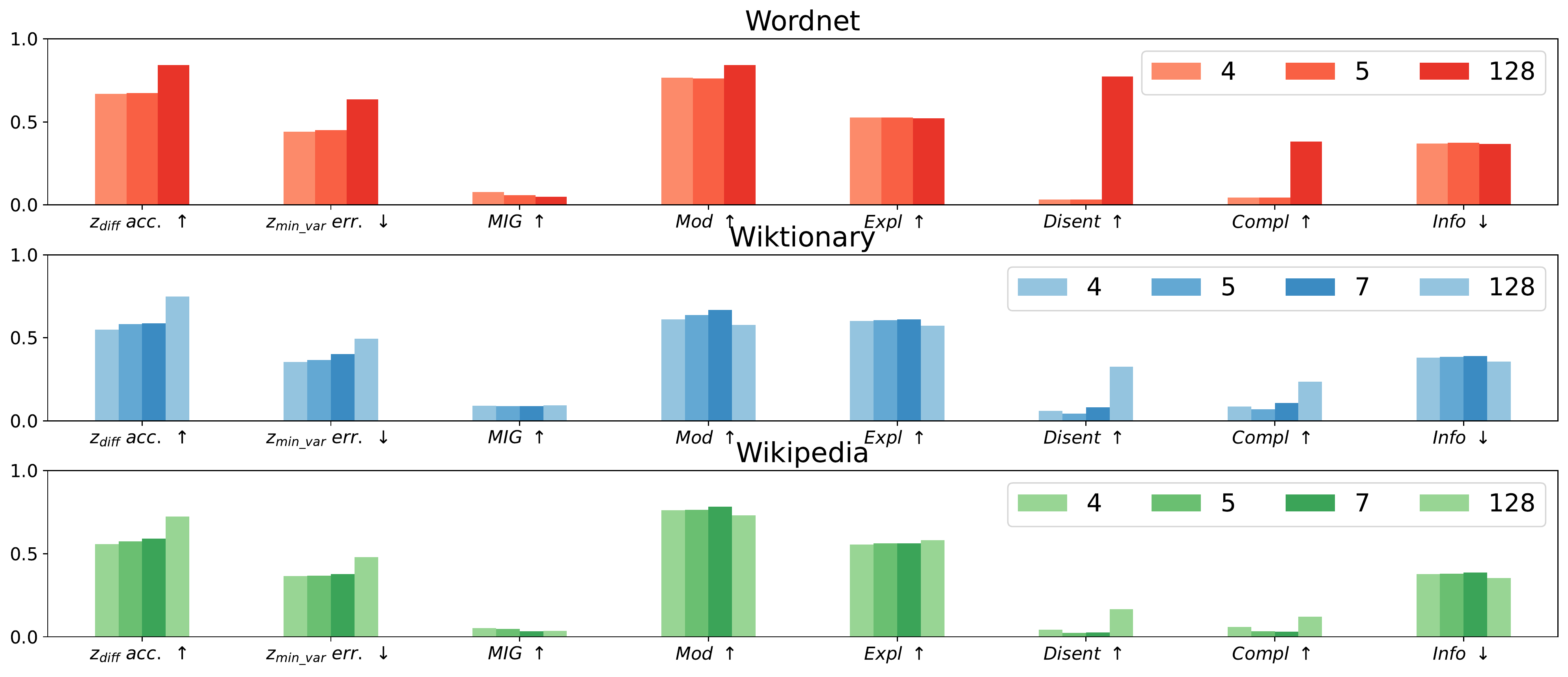}
        \caption{By number of latents}
        \label{fig:all_metrics_by_groups}
    \end{subfigure}
    \caption{Metrics mean grouped.}
\end{figure*}

\input{sections/quantitative.tex}

\input{sections/generation.tex}

%% file: sections/exp_setup.tex
\subsection{Experimental setup}

\noindent
\textbf{Datasets}
Definition sentences and their respective semantic role structures are sourced from three different datasets by~\citep{silva2016categorization} with the characteristics described in Table~\ref{tab:stats_data}. All datasets are automatically annotated with DSR tags for each token, using the method proposed by~\citep{silva2016categorization}. The datasets differ not only in sentence length and size, but also in textual style: while WordNet and Wiktionary sentences tend to be formatted as dictionary definitions, Wikipedia sentences are lengthier and less adherent to a typical definition structure. For brevity, hyperparameter choices and implementation details are covered in sections \ref{seq:hyperparams} and \ref{seq:impl_details} of the suplementary material.

\input{tables/stats_data.tex}

%% file: tables/stats_data.tex
\begin{table}[t]
    \small
    \centering
    \begin{tabular}{|l|ccc|}
        \hline
        Dataset & Num sents. & Avg. length & Version \\ \hline
        Wordnet & 93,699 & 9 & WordNet 3.0 \\
        Wiktionary & 464,243 & 8 & Dec, 2016 \\
        Wikipedia & 1,500,323 & 12 & Dec, 2016 \\
        \hline
    \end{tabular}
    \caption{Statistics from definition datasets.} \label{tab:stats_data}
\end{table}

%% file: sections/qualitative.tex
\subsection{Qualitative Evaluation}
We analyse the representations of the trained models in terms of their disentanglement and composition, by applying three different techniques 1) traversals of the latent space, 2) latent space arithmetic, 3) encoding interpolation.\\ 

\noindent
\textbf{Latent space traversals}
Traversal evaluation is a standard procedure with image disentanglement~\citep{higgins2016beta,kim2018disentangling}. The traversal of a latent factor is obtained as the decoding of the vectors corresponding to the latent variables, where the evaluated factor is changed within a fixed interval, while all others are kept fixed. If the representation is disentangled, when a latent factor is traversed, the decoded sentences should only change with respect to that factor. This means that after training the model we are able to probe the representation for each latent variable. In the experiment, the traversal is set up from a starting point given by a ``seed'' sentence. As illustrated in Table~\ref{tab:trav_examples} we observed that the latent variables typically track a single abstract definition role (e.g., supertype, quality, purpose), and change the meaning of the original term according to an abstract interpretation axis (e.g, flying $\rightarrow$ \textit{movement}, art $\rightarrow$ \textit{doutrine/teachings}). This means a certain degree of control can be applied to the generation of both the sentence structure and semantics.\\ 

\input{tables/trav_examples.tex}

\noindent
\textbf{Latent space arithmetic}
In this experiment, the latent vectors for two sentences are added, subtracted or averaged, and then the resulting vectors are traversed. The sentence pairs are different by a single term, so that we can observe the latent variables affected by the change, and how they are affected. As illustrated in Table~\ref{tab:arith_examples}, these operations tend to produce vectors that, when traversed, generate sentences corresponding to the features manipulated by the operation (e.g., removing the \textit{monarch} supertype, leaving the \textit{female} quality).\\

\input{tables/arith_examples.tex}

\noindent
\textbf{Interpolation}
In this experiment, we analyse the capability of the models built with the proposed approach to provide a smooth transition between latent space representations of sentences~\citep{bowman2016generating}. In practice, the interpolation mechanism takes two sentences $ x_1 $ and $ x_2 $, and uses their posterior mean as the latent features $ z_1 $ and $ z_2 $, respectively. It interpolates a path $ z_t = z_1 \cdot (1 - t) + z_2 \cdot t  $  with $ t $ increased from $ 0 $ to $ 1 $ by a step size of $ 0.1 $.  This is a deterministic process, and no search is performed. As a result, $ 9  $ sentences are generated on each interpolation step. In Table~\ref{tab:interpol_examples} we provide qualitative results with latent space interpolation on Wiktionary. We can observe the transition happening for each concept: \textit{migratory} $\rightarrow$ $\emptyset$ $\rightarrow$ \textit{microscopic}, \textit{aquatic} $\rightarrow$ \textit{aquatic + terrestrial} $\rightarrow$ \textit{terrestrial}, \textit{bird} $\rightarrow$ \textit{mammal} $\rightarrow$ \textit{organism} $\rightarrow$ \textit{invertebrate}. This type of localised semantic control provided by the operations of traversal and interpolation over intensional-level (definitional) sentences can potentially support quasi-symbolic operations over the latent space. Such effects could not be observed within the baselines.

\input{tables/interpol_examples.tex}

\begin{figure}[ht]
    \centering
    \includegraphics[scale=0.52]{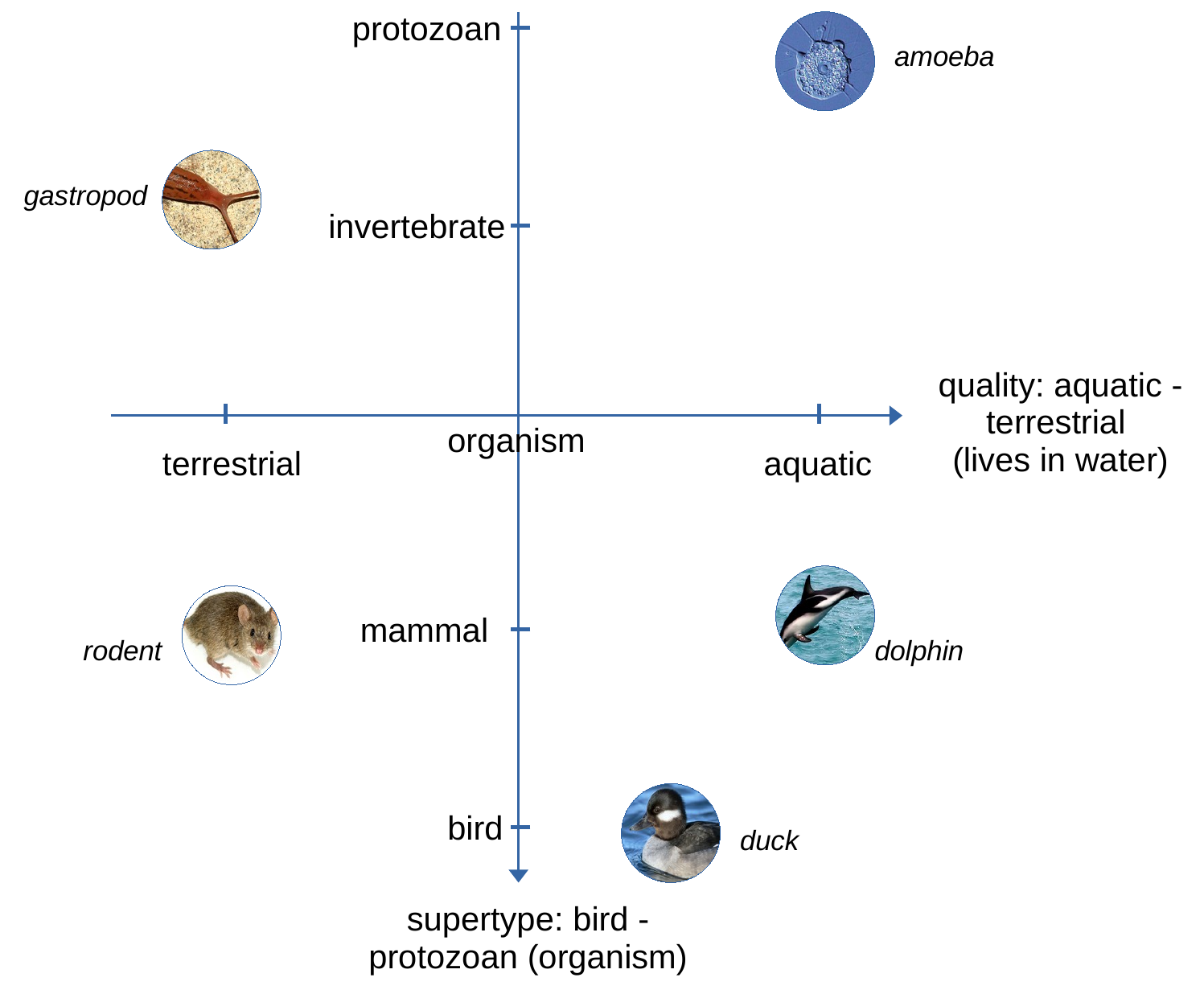}
    \caption{Conceptualisation of a two-dimension cut of the latent space, applied to the first example in Table~\ref{tab:interpol_examples}.}
    \label{fig:latent_concept}
\end{figure}

Based on those three experiments, the composition of such latent space could be conceptualised as in the projection illustrated in Figure~\ref{fig:latent_concept}.

\begin{figure}[t]
    \includegraphics[scale=0.23]{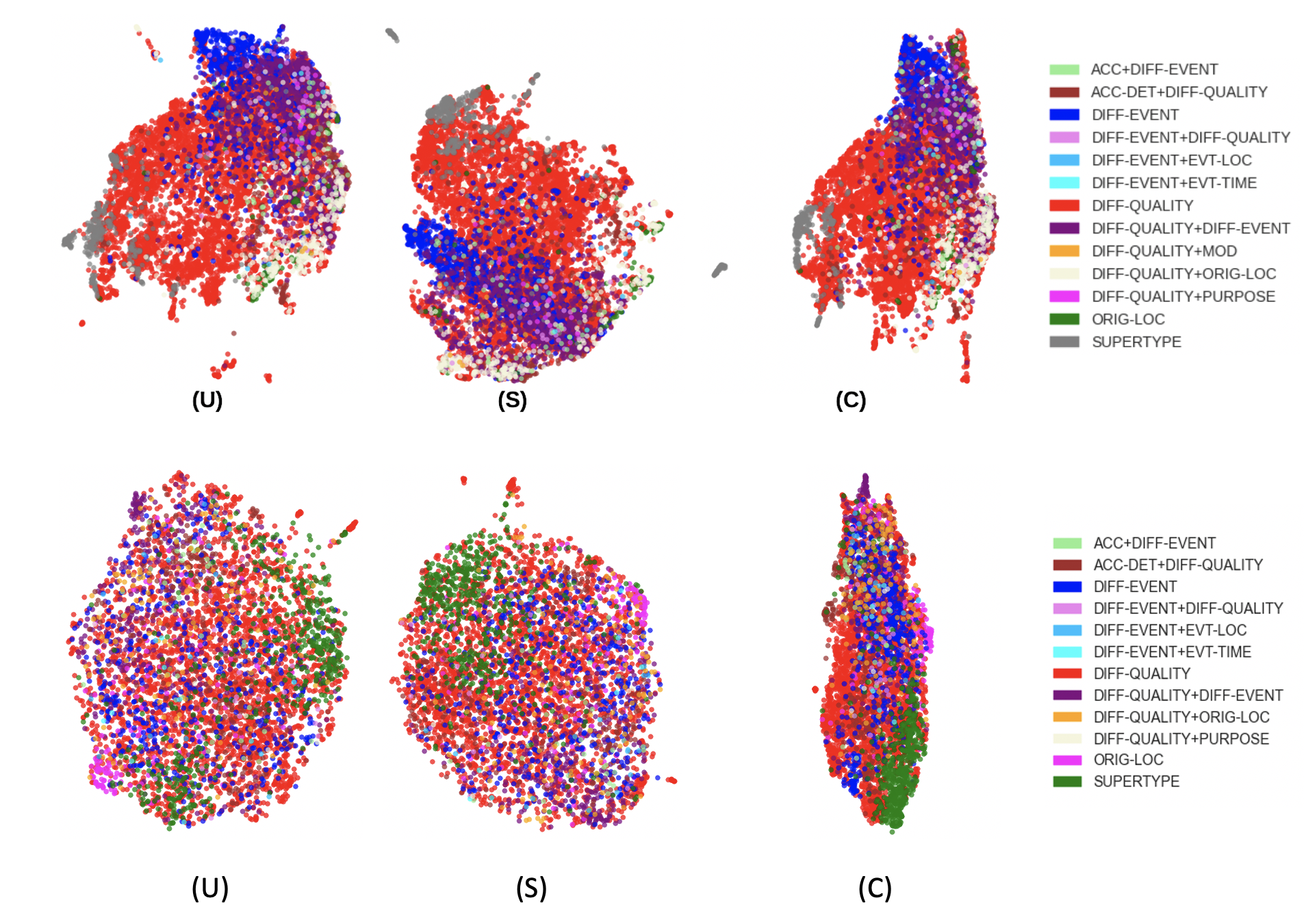}
    \caption{UMAP plot of latent representations from Unsupervised VAE (U), DSR supervision (S) and Conditional VAE (C) (Top: LSTM, Bottom: Optimus-based).}
    \label{fig:umap}
\end{figure}

\paragraph{UMAP plot}
\textit{UMAP} (Uniform Manifold Approximation and Projection)~\cite{mcinnes2018umap} is a popular method for non-linear dimensionality reduction, that allows the visualization of complex high-dimensional feature spaces, such as the representation space produced by a VAE. Figure~\ref{fig:umap} presents a 2D plot of UMAP transformations for both baselines under three training frameworks, from which the clustering of DSR patterns can be observed. While the supervision with DSR labels promotes clustering of the patterns around the center of the plot, cVAE compacts the cluster on the edges, allowing better separation. In the Optimus-based model, for example, the \textit{SUPERTYPE} (green) cluster has a tendency to move towards the edge of plot from left (U) to right (C). t-SNE transformations are also performed and the plots are presented in the supplemental material (Appendix~\ref{sec:apd_exp_res}).

%% file: tables/trav_examples.tex

%
%

\begin{table}[t]
\scriptsize

\begin{tikzpicture}
\node (table) [inner sep=.1pt] {
\begin{tabular}{l}
\setlength\tabcolsep{.2pt}
\underline{a flying creature} \\
\textcolor{blue}{a flying} \textcolor{red}{animal}  \\
\textcolor{blue}{a flying} \textcolor{red}{insect} \\

~\\

a \textcolor{red}{robot} \\
a \textcolor{red}{monster} \\
a \textcolor{red}{creature} \\

~\\

\textcolor{red}{a walking} \textcolor{blue}{demon} \\
\textcolor{red}{a flying} \textcolor{blue}{creature} \\
\textcolor{red}{a moving} \textcolor{blue}{animal} \\

\end{tabular}
};
\draw [rounded corners=.5em, line width=.5pt] (table.north west) rectangle (table.south east);
\end{tikzpicture}
\begin{tikzpicture}
    \node (table) [inner sep=.1pt] {
    \begin{tabular}{l l}

\multirow{3}{*}{\rotatebox[origin=c]{90}{}}
& \underline{a martial art developed in Israel} \\
& \textcolor{blue}{an ancient Buddhist dagger} \textcolor{blue}{used to stab others}  \\
& \textcolor{blue}{an ancient martial art} \textcolor{blue}{practiced in Japan} \\
& \\

\multirow{3}{*}{\rotatebox[origin=c]{90}{}}
& \textcolor{blue}{a Roman soldier's} \textcolor{red}{movement} \\
& \textcolor{blue}{a military} \textcolor{red}{dress worn by monks} \\
& \textcolor{blue}{a knight's} \textcolor{red}{ceremonial hat} \\

& \\
\multirow{3}{*}{\rotatebox{90}{}}
& \textcolor{red}{a religious rite} in which communion is offered \\
& \textcolor{red}{a literary rite} in Bible study \\
& \textcolor{red}{a medicine school} \\

\end{tabular}
};
\draw [rounded corners=.5em, line width=.5pt] (table.north west) rectangle (table.south east);
\end{tikzpicture}

\caption{Traversals showing \textcolor{red}{changed} and \textcolor{blue}{held} semantic factors in Wiktionary definitions (Optimus-based model).}
\label{tab:trav_examples}
\end{table}


%% file: tables/arith_examples.tex

%
%

\begin{table}[t]
\scriptsize

\begin{center}
\begin{tikzpicture}
\node (table) [inner sep=.1pt] {
\begin{tabular}{l l}
\setlength\tabcolsep{.5pt}
\multirow{5}{*}{\rotatebox[origin=c]{90}{ADD}}
& \underline{a flying machine}~~~ \\
& \underline{a flying creature} \\
& \textcolor{blue}{a flying} \textcolor{red}{dinosaur}  \\
& \textcolor{blue}{a flying} \textcolor{red}{robot} \\
& \textcolor{blue}{a flying} \textcolor{red}{object} \\

\end{tabular}
};
\draw [rounded corners=.5em, line width=.5pt] (table.north west) rectangle (table.south east);
\end{tikzpicture}
\begin{tikzpicture}
\node (table) [inner sep=.1pt] {
\begin{tabular}{l l}
\setlength\tabcolsep{.5pt}
\multirow{5}{*}{\rotatebox[origin=c]{90}{AVG}}
& \underline{to make four copies of~~} \\
& \underline{to make five copies of} \\
& \textcolor{blue}{to make} \textcolor{red}{one} \textcolor{blue}{copy of}  \\
& \textcolor{blue}{to make} \textcolor{red}{two} \textcolor{blue}{copies of} \\
& \textcolor{blue}{to make} \textcolor{red}{3} \textcolor{blue}{copies of} \\

\end{tabular}
};
\draw [rounded corners=.5em, line width=.5pt] (table.north west) rectangle (table.south east);
\end{tikzpicture}

\begin{tikzpicture}
\node (table) [inner sep=.1pt] {
\begin{tabular}{l l}
\setlength\tabcolsep{.5pt}
\multirow{5}{*}{\rotatebox[origin=c]{90}{SUB}}
& \underline{a female monarch} \\
& \underline{a monarch} \\
& \textcolor{red}{the subnormal condition in} \textcolor{blue}{females}  \textcolor{red}{originating from...} \\
& \textcolor{red}{the normal} \textcolor{blue}{female pregnancy}  \textcolor{red}{associated with some} \\
& \textcolor{red}{the} \textcolor{blue}{female}  \textcolor{red}{given name in the Japanese game...} \\
\end{tabular}
};
\draw [rounded corners=.5em, line width=.5pt] (table.north west) rectangle (table.south east);
\end{tikzpicture}

\end{center}

\caption{Traversals showing \textcolor{red}{changed} and \textcolor{blue}{held} semantic factors after latent vector arithmetic in Wiktionary definitions (Optimus-based model).}
\label{tab:arith_examples}
\end{table}


%% file: tables/interpol_examples.tex
%
%

\begin{table}[t]
\scriptsize

\begin{center}
\begin{tikzpicture}
\node (table) [inner sep=.1pt] {
\begin{tabular}{ l l}
\setlength\tabcolsep{.5pt}



\multirow{12}{*}{\rotatebox[origin=c]{90}{DSR Optimus-based}}
& \textcolor{blue}{a migratory aquatic bird found in the temperate regions} \\
& \textcolor{blue}{of the northern hemisphere} \\
& 1 a migratory bird of the eastern Mediterranean \\
& 2 a marine gastropod of the subfamily \\
& 3 a terrestrial aquatic mammal of the family \\ 
& 4 a terrestrial aquatic mammal of the suborder \\
& 5 a terrestrial invertebrate \\
& 6 a microscopic organism or invertebrate \\
& \textcolor{blue}{a microscopic terrestrial animal or protozoan} \\

& \\

& \textcolor{blue}{an automobile} \\
& 1 a motorcycle \\
& \textcolor{blue}{a bicycle} \\

\end{tabular}
};
\draw [rounded corners=.5em, line width=.5pt] (table.north west) rectangle (table.south east);
\end{tikzpicture}
\end{center}

\caption{Interpolation examples in Wiktionary (Optimus-based model). Only unique sentences are shown.}
\label{tab:interpol_examples}
\end{table}


%% file: sections/quantitative.tex
\subsection{Quantitative Evaluation}
In this experiment we probe the representation learned by the proposed VAE models using eight popular quantitative metrics for disentanglement, namely: z-diff~\citep{higgins2016beta}, z-min-var~\citep{kim2018disentangling}, Mutual Information Gap (MIG)~\citep{chen2018isolating}, Modularity \& Explicitness~\citep{ridgeway2018learning}, and from~\citep{eastwood2018framework}(disentanglement, completeness, informativeness). Further details about the metrics are provided in Appendix~\ref{sec:metrics}. It is relevant to mention that there are considerations regarding inconsistency on classification dependent probes (e.g., z-min-var, modularity), which are not discussed here due to space and scope considerations (we refer to \citet{carbonneau2022measuring}). Therefore, we decided to include all current metrics that could be applied in this scenario, and the results presented next should be interpreted considering these limitations.\\

\noindent
\textbf{Experimental Setup}
We evaluate VAE (U), DSR VAE (S) and CVAE (C) on Wordnet (WN), Wiktionary (WT) and Wikipedia (WP) datasets. Evaluation is performed under the framework explained in Section~\ref{sec:eval}. Each combination of VAE architecture, generative factor grouping and representation size was trained and quantitatively tested, by calculating the previously mentioned disentanglement metrics. For computing the metrics we follow the experiments of~\citet{zhang2021unsupervised}.

\noindent
\textbf{Analysis}
The results presented in Tables~\ref{tab:trav_examples},~\ref{tab:interpol_examples}, and ~\ref{tab:all_metrics} show that, specially when using the Optimus-based model:

\input{tables/all_ds_by_exp.tex}

\input{tables/generation_examples.tex}

\textit{For the Wiktionary and Wikipedia datasets, the application of DSR categories as biases results in a measurable improvement in disentanglement} (RQ1).
This is evidenced by the proposed model outperforming the unsupervised baseline in six of the eight disentanglement metrics tested, by a margin of at least 2.5\%, 81\% in average.

\textit{The use of DSRs as generative factors produces meaningful disentangled representations} (RQ2). The traversal results indicate the tendency of associating certain role abstractions to latent space dimensions, e.g., supertype,  statement (purpose, among others). The interpolation results indicate the capture of semantic bridging across definitions, e.g., teaching $\rightarrow$ loading (process). The UMAP visualisation indicates slightly better factor separation and smoother transitions for the conditional model.
  
More specifically, in LSTM, z-diff presents the highest and most consistent improvement, specially with the CVAE, indicating higher interpretability when inferring single generative factors from the representations. Explicitness results are also consistent, indicating higher coverage of each factor. Improvements on Modularity, Disentanglement Score, Completeness and Informativeness are less consistent, indicating that the factors share substantial information between them. On the other hand, z-min-var, MIG counter the trend of improvement, due to the fact that they are designed to strongly penalize non-alignment of single pairs <factor $\leftrightarrow$ latent dimension> (e.g., linear combinations). As a result, they penalize the existence of dependency and hierarchy relations which is present in most DSR categories, e.g., DIFFERENTIA-EVENT $\rightarrow$ EVENT-TIME. As for the Optimus-based model, there are similar tendencies on WT and WP corpus. The conditional framework always performs better under 6 of 8 metrics, except z-min-var and modularity. This result indicates that our conditional framework can improve the disentanglement performance of Optimus.

We also analyse how semantic groupings affect disentanglement in Figure~\ref{fig:all_metrics_by_groups} (RQ3). This is done only for the LSTM-based VAE, as the Transformer-based one was set to the optimal configuration in \citet{li2020optimus}. Overall, we notice that syntax based groups have higher scores, indicating that it is easier to disentangle syntactic phrase components. For Modularity the result is the opposite, indicating that semantic groupings promote higher independence between factors. Following~\citep{zhang2021unsupervised}, the values in Table~\ref{tab:all_metrics} for the metrics Completeness and Disentanglement score are multiplied by 10, in order to facilitate the visualization.

Finally, we find that a low number of latent dimensions leads to smaller degree of disentanglement. The experiments with 4,5,7 and 128 latents are reported in Figure~\ref{fig:zdim_analysis}.



%% file: tables/all_ds_by_exp.tex

\begin{center}
\begin{table}[ht!]
\scriptsize
\setlength\tabcolsep{2.5pt}
\resizebox{7.8cm}{!}{
\begin{tabular}{ c |ccc|ccc|ccc|ccc }
\hline
\multicolumn{12}{c}{LSTM} \\ \hline
D & \multicolumn{3}{c}{z-diff} & \multicolumn{3}{c}{z-min-var $\downarrow$} & \multicolumn{3}{c}{MIG} & \multicolumn{3}{c}{Modularity}   \\ \hline
  & U    & S             & C             & U             & S             & C             & U             & S             & C             & U             & S             & C  \\ \hline
WN & .700 & .691          & \textbf{.770} & \textbf{.482} & .503          & .532          & \textbf{.067} & .057          & .059          & .793          & \textbf{.804} & .765  \\
WT & .597 & .619          & \textbf{.635} & .400          & \textbf{.385} & .430         & \textbf{.112} & .095          & .065          & .535          & .424 & \textbf{.629}   \\
WP & .575 & .630          & \textbf{.647} & .398          & \textbf{.386} & .420          & \textbf{.046} & .041          & .037          & \textbf{.771} & .745 & .757  \\ \hline

D & \multicolumn{3}{c}{Explicitness}  & \multicolumn{3}{c}{Disentanglement}  & \multicolumn{3}{c}{Completeness}  & \multicolumn{3}{c}{Informativeness $\downarrow$} \\ \hline
  & U    & S             & C             & U             & S             & C             & U             & S             & C             & U             & S             & C   \\ \hline
WN & .519 & \textbf{.532} & .527          & .022          & .021          & \textbf{.031} & .013          & .013          & \textbf{.017} & .364          & \textbf{.361} & .399 \\
WT & .584 & .593          & \textbf{.616} & \textbf{.014} & .011          & .013         & \textbf{.013} & \textbf{.013} & .011 & .377 & \textbf{.373} & .385    \\
WP & .545 & .557          & \textbf{.600} & \textbf{.007} & \textbf{.007} & .005          & \textbf{.007} & \textbf{.007} & .004& .375 & \textbf{.373} & .374 \\ \hline 
\multicolumn{12}{c}{Optimus-based} \\ \hline
D & \multicolumn{3}{c}{z-diff} & \multicolumn{3}{c}{z-min-var $\downarrow$} & \multicolumn{3}{c}{MIG} & \multicolumn{3}{c}{Modularity}   \\ \hline
  & U    & S             & C             & U             & S             & C             & U             & S             & C             & U             & S             & C  \\ \hline
WN & .645 & \textbf{.673} & .669    & \textbf{.483} & .509  & .517          & \textbf{.023} & .012          & .006          & .724         & \textbf{.766}  & .750  \\
WT & .516 & .532          & \textbf{.589} & .458  & \textbf{.441} & .480         & .016& .013          & \textbf{.043}           & \textbf{.827}         & .813     & .809   \\
WP & .513 & .544          & \textbf{.641} & \textbf{.471}  & .486 & .552         & .010 & .011          & \textbf{.033}                       & \textbf{.956}       & .942     & .943  \\ \hline

D & \multicolumn{3}{c}{Explicitness}  & \multicolumn{3}{c}{Disentanglement}  & \multicolumn{3}{c}{Completeness}  & \multicolumn{3}{c}{Informativeness $\downarrow$} \\ \hline
  & U    & S             & C             & U             & S             & C             & U             & S             & C             & U             & S             & C   \\ \hline
WN & .501 & .500 & \textbf{.501}          & \textbf{.058} & .040 & .049     & \textbf{.039} & .027  & .032      & .398  & \textbf{.377} & .398 \\
WT & .559 & .547  & \textbf{.573}         & .013 & .026 & \textbf{.028}      & .009 & .018 & \textbf{.019}     & .333 & .316 & \textbf{.305}    \\
WP & .548 & .532        & \textbf{.594}              & .024 & .054 & \textbf{.060}                 & .016 & .034& \textbf{.038}                      & .288 & .282 & \textbf{.280} \\ \hline

\end{tabular}
}
\caption{Quantitative disentanglement metrics (Top: LSTM, Bottom: Optimus-based). } \label{tab:all_metrics}
\end{table}
\end{center}


%% file: tables/generation_examples.tex
%
%

\begin{table*}[ht!]
\scriptsize

\begin{center}
\begin{tikzpicture}
\node (table) [inner sep=.1pt] {

    \begin{tabular}{llll}
        \setlength\tabcolsep{.5pt}
        Word          & Definition Model                       & Unsupervised LSTM                                  & Supervised LSTM                                \\ \hline
        repulse       & the act of making a gun                & the act of moving forward                         & act in a hostile state                        \\
        colonise      & make a new or vital part               & the state of being in a particular place          & settle or cause to be easily removed          \\
        involve       & make a specific purpose                & make a specific effect                            & a specific act of making something            \\
        mitochondrion & a cell that is used to treat the blood & a substance that is used to treat a body reaction & a cell that is a source of an organic process \\
        heat          & a change in the surface of a liquid    & a sudden increase in the flow of heat             & a sudden increase in the temperature          \\

    \end{tabular}
};
\draw [rounded corners=.5em, line width=.5pt] (table.north west) rectangle (table.south east);
\end{tikzpicture}
\end{center}

\caption{Definition generation examples for the Wordnet dataset.}
\label{tab:generation_examples}
\end{table*}


%% file: sections/generation.tex
\subsection{Definition Generation}
In this experiment, we assess the proposed VAE models in the task of "Definition Modeling"~\citep{noraset2017definition}, where the goal is to generate a natural language definition given the word to be defined (definiendum).\\

\noindent
\textbf{Experimental setup}
During training, we adopt the "seed" setup~\citep{noraset2017definition}, which involves providing the definiendum concatenated with the definition tokens as input for the model. At generation time, the model takes as input only the word which needs to be defined, and leverages a trained model for computing the definition latent encoding. Such encoding is then fed into a softmax function and subsequently a multinomial probability distribution is sampled for decoding the latent variable into the final definition sentence.

To compare with the baseline of definition generation \citep{gadetsky2018conditional}, we only consider LSTM-based VAEs under the proposed unsupervised and DSR-supervised framework, both using the "seed" setup. The conditional LSTM and optimus-based models are not explored in this experiment in order to have a more fair comparison with the Definition model. We train the baseline and our models with similar setups, following~\citep{gadetsky2018conditional}. We perform language model pretraining on the WikiText-103 dataset~\citep{merity2016pointer} for 1 epoch, then train on the downstream dataset for 10 epochs. Additionally, all models are initialised using Google Word2Vec pretrained vectors, following~\citep{gadetsky2018conditional}.\\

\noindent
\textbf{Results}
We report the perplexity and Bleu~\citep{papineni2002bleu} results in Table~\ref{tab:results_generation}. We observe that the proposed variational autoencoder models achieve an improvement on both perplexity and Bleu compared to the RNN baseline. The DSR VAE achieves the best perplexity and Bleu on 2 out of 3 datasets while the unsupervised VAE is the best performing model in the other cases. Success of VAE models can be attributed to their disentangling properties, which promotes learning of latent spaces that are less sparse, a benefit deriving from sampling variable for re-parameterization. Improvements from the DSR VAE are marginal, but can be attributed to the additional information that is injected into its latent variables.

\input{tables/generation_results.tex}

Some generation examples from the Wordnet dataset are provided in Table~\ref{tab:generation_examples}. Such examples show that the proposed VAE models are able to leverage the structural and semantic information of the learned definition roles to better approximate the defined concept. In particular, we notice some semantically strong linguistic elements in the definitions decoded with DSR supervision, for example DSR is the only model able to link the verb "repulse" with the hostile adjective, the verb colonise with the similar verb "settle", and the word "heat" with temperature. We include more generation examples of the Optimus-based model in Appendix~\ref{sec:apd_exp_res}.

The strong performance in this definition generation task indicates that the disentangled representations have provided the VAE models with higher generalization capability, suggesting that disentangling is beneficial for diverse applications.

%% file: tables/generation_results.tex
\begin{table}[ht!]
    \small
    \centering
    \begin{tabular}{lccc|ccc}
        \hline
        & \multicolumn{3}{c}{Perplexity $\downarrow$} & \multicolumn{3}{c}{Bleu} \\ \hline
        Data & DM    & VAE            & DSR            & DM    & VAE            & DSR            \\ \hline
        WN   & 88.59 & 80.36          & \textbf{80.27} & 9.12  & \textbf{10.27} & 10.26          \\
        WT   & 42.51 & 39.09          & \textbf{38.64} & 6.70  & 7.53           & \textbf{7.59}  \\
        WP   & 13.09 & \textbf{12.39} & 12.47          & 11.89 & 12.32          & \textbf{12.34} \\
        \hline
    \end{tabular}
    \caption{Quantitative metrics for definition generation.} \label{tab:results_generation}
\end{table}

%% file: sections/conclusion.tex
\section{Conclusion}
We propose a novel VAE-based framework for learning and evaluating disentangled representations in natural language definitions. We leverage the semantic structure present in dictionaries as inductive biases for improving disentanglement in VAEs, and as generative factors during evaluation. Our evaluation shows, both with qualitative investigations and with quantitative metrics, that the proposed framework is able to produce encodings with a higher degree of disentanglement. Finally, our models outperform existing baselines on a definition modeling application, demonstrating the generalization capabilities of disentangled representations.


%% file: sections_eacl_2023/limitations.tex
\section*{Limitations}
The type of structural supervision chosen for the approach here proposed is specificaly fit to definition (dictionary style) sentences, in order to leverage semantic information from such structures. However, this limits the scope of comparison with other methods applied to general sentences. Additionally, the qualitative improvements we observed in terms of latent space traversals, arithmetic and interpolation do not clearly correlate with the disentanglement metrics, despite overall improvement. This raises some questions regarding the relation between explainability properties and general latent space separability.

%% file: sections/appendix.tex
\clearpage

\appendix

\section{Definition Semantic Roles} \label{sec:dsr_labels}
The datasets used in our experiments are introduced in~\citep{silva2018recognizing}. We report in Table~\ref{tab:srl_silva} the annotated categories.
\input{tables/silva_srl.tex}


\input{sections/metrics.tex}


\section{Hyperparameter choices} \label{seq:hyperparams}
Experiments are conducted to cover a set of 3 hyperparameters: First, the VAE architecture used: 1) Unsupervised VAE 2) Supervised with SRL 3) CVAE with SRL. Second, the generative factor grouping, which includes: 1) Semantic w/ supertype 2) Syntactic w/ supertype 3) Semantic w/o supertype 4) Syntactic w/o supertype. Third, the dimensionality of VAE latent representation ($z$): 4, 5, 7, 128.

The choice of architecture allows evaluation of the impact of DSR label conditioning in two distinct ways: as part of the autoencoding objective function, and as a conditional variable of the decoder, addressing our research questions \textbf{RQ1} and \textbf{RQ2}. The choice of generative factor grouping can indicate the best ways to organize the factors, addressing \textbf{RQ3}.

The dimensionality of the representation is set to match the number of generative factors, in an attempt to force disentanglement by alignment of each dimension to a single factor. The dimension sizes are then defined to be 4 (alignment with groupings 3 and 4), 5 (alignment with grouping 2) or 7 (alignment with grouping 1). However, different levels of disentanglement can be achieved with mismatching dimensions and factors. So all possible combinations of factors and representation sizes are tested and a size of 128 is included to evaluate the impact of a higher number of parameters in each grouping.

\section{Implementation Details} \label{seq:impl_details}
As for LSTM-based VAE, hyperparameters are chosen with the following values, based on a previous experiment from~\citep{shen2020educating}. (1) Number of hidden layers: 1, (2) Dimension of the hidden layer: 512, (3) VAE $\lambda_{KL}=0.1$, (4) Epochs=20, (5) Batch size=32 for Wikipedia, 64 for the rest. Dropout (20\%) is done for both encoder and decoder inputs. To provide the inputs and outputs for the VAEs, the definition sentences are tokenized into sub-words with a \textit{Byte Pair Encoding} (BPE) scheme, and converted into token embeddings with the T5 transformer model~\citep{Raffel2020t5}, with an embedding size of 512. With respect to Optimus, we use memory setup to inject latent representation into the decoder. The encoder and decoder are pretrained BERT with bert-base-cased version and GPT2, respectively. Some additional values of hyperparameters are:  (1) Epochs=10, (2) Batch size=32. (3) latent size=32. In the supervised framework, a new embedding layer is considered to learn the representations of semantic roles. In the conditional framework, we add semantic roles into the vocabulary of pretrained BERT encoder.


\section{Further Experimental Results} \label{sec:apd_exp_res}
\begin{figure}[h!]
    \includegraphics[scale=0.29]{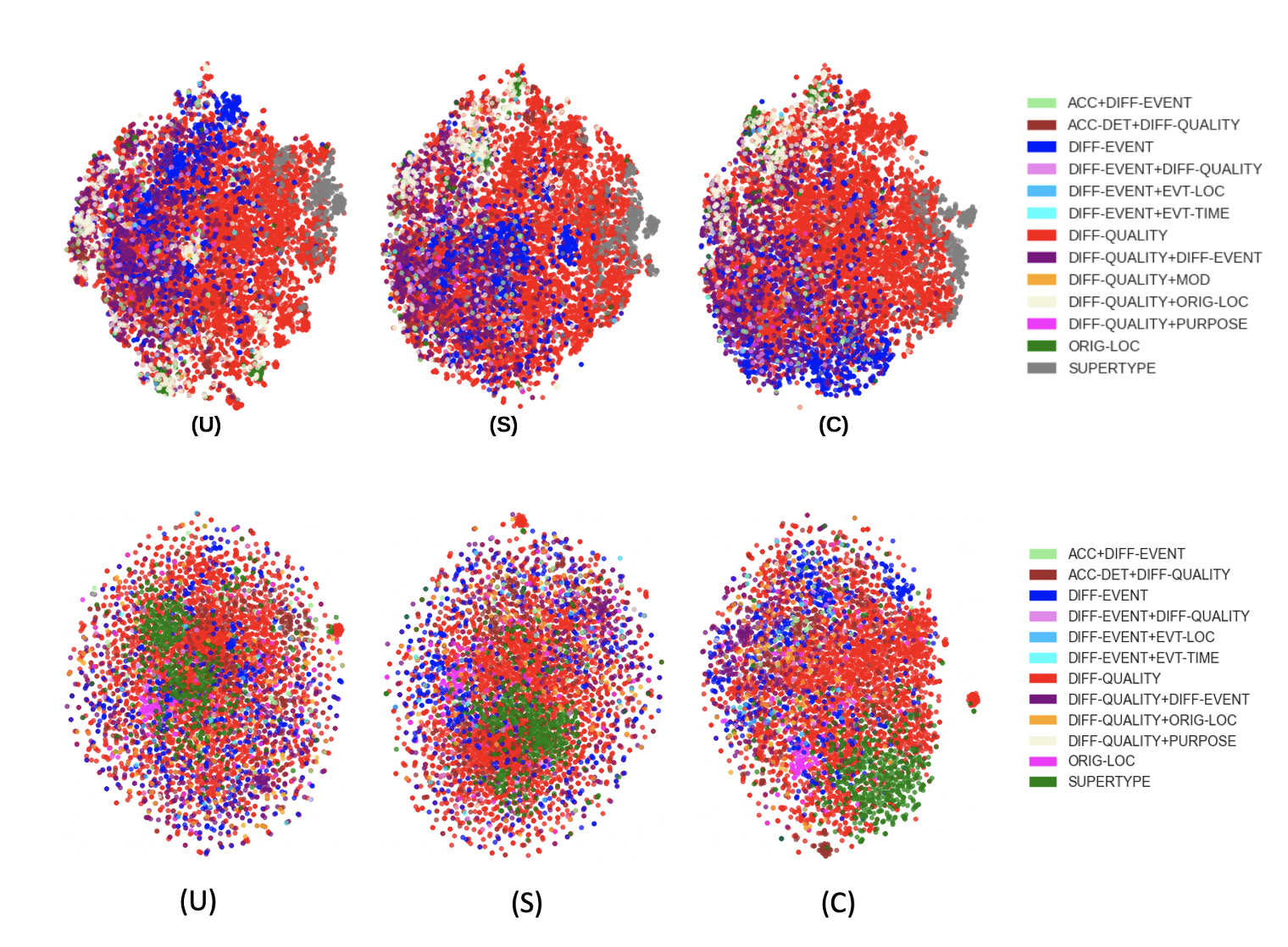}
    \caption{t-SNE plot of latent representation generated from LSTM and Optimus-based models under Unsupervised VAE (U), DSR supervision (S) and Conditional VAE (C) (Top: LSTM, Bottom: Optimus-based).}
    \label{fig:tsne}
\end{figure}

\paragraph{t-SNE plot}
Alternative dimensionality reduction method (t-distributed Stochastic Neighbor Embedding)~\cite{van2008visualizing}, used to visualise the clustering of DSR patterns, as seen in Figure~\ref{fig:tsne}. 

\paragraph{Optimus-based model definition generation} 
Table~\ref{tab:srl_silva} lists the generated definitions from the Unsupervised Optimus-based model on Wordnet. The perplexity is 35.46 that is much lower than 80.27 from LSTM. 
\input{tables/optimus_def_examples.tex}


%
%
%
%

%% file: tables/silva_srl.tex
\begin{table}[h]
    \small \setlength\tabcolsep{4.5pt}
    \centering
    \begin{tabular}{|p{1.5cm}|p{6cm}|}  \hline
        \textbf{Role}         & \textbf{Description}                       \\ \hline
        Supertype & the immediate or ancestral entity’s superclass \\ \hline
        Differentia quality & a quality that distinguishes the entity from the others under the same supertype \\ \hline
        Differentia event & an event (action, state or process) in which the entity participates and that is mandatory to distinguish it from the others under the same supertype \\ \hline
        Event \qquad location & the location of a differentia event \\ \hline
        Event time & the time in which a differentia event happens \\ \hline
        Origin \qquad location & the entity’s location of origin \\ \hline
        Quality modifier & degree, frequency or manner modifiers that constrain a differentia quality \\ \hline
        Purpose & the main goal of the entity’s existence or occurrence \\ \hline
        Associated fact & a fact whose occurrence is/was linked to the entity’s existence or occurrence \\ \hline
        Accessory determiner & a determiner expression that doesn’t constrain the supertype / differentia scope \\ \hline
        Accessory quality & a quality that is not essential to characterize the entity \\ \hline
        Role \qquad particle & a particle, such as a phrasal verb complement, non-contiguous to the other role components \\ \hline
    \end{tabular}
    \caption{Semantic Role Labels for dictionary definitions.} 
    \label{tab:srl_silva}
\end{table}

%% file: sections/metrics.tex
\section{Disentanglement Metrics} \label{sec:metrics}

\begin{enumerate}
    \item $z_{diff}~accuracy$~\cite{higgins2016beta}: The accuracy of a predictor for $p(y|z^b_{diff})$, where $z^b_{diff}$ is the absolute linear difference between the inferred latent representations for a batch $B$ of latent vectors, written as a percentage value. Higher values imply better disentanglement.

    \item $z_{min\_var}~error$~\cite{kim2018disentangling}: For a chosen factor k, data is generated with this factor fixed but all other factors varying randomly; their representations are obtained, with each dimension normalised by its empirical standard deviation over the full data (or a large enough random subset); the empirical variance is taken for each dimension of these normalised representations. Then the index of the dimension with the lowest variance and the target index k provide one training input/output example for the classifier. Thus, if the representation is perfectly disentangled, the empirical variance in the dimension corresponding to the fixed factor will be 0. The representations are normalised so that the arg min
    is invariant to rescaling of the representations in each dimension. Since both inputs and outputs lie in a discrete space, the optimal classifier is the majority-vote classifier, and the metric is the error rate of the classifier. Lower values imply better disentanglement.

    \item Mutual Information Gap ($MIG$)~\cite{chen2018isolating}: The difference between the top two latent variables with the highest mutual information. Empirical mutual information between a latent representation $z_j$ and a ground truth factor $v_k$, is estimated using the joint distribution defined by  $q(z_j, v_k) = \sum_{n=1}^{N}{p(v_k)p(n|v_k)q(z_j|n)}$. A higher mutual information implies that $z_j$ contains a more information about $v_k$, and the mutual information is maximal if there exists a deterministic, invertible relationship between $z_j$ and $v_k$. MIG values are in the interval [0, 1], with higher values implying better disentanglement.

    \item \textit{Modularity}~\cite{ridgeway2018learning}: The deviation from an ideally modular case of latent representation. If latent vector dimension $i$ is ideally modular, it will have high mutual information with a single factor and zero mutual information with all other factors. A deviation $\delta_i$ of 0 indicates perfect modularity and 1 indicates that this dimension has equal mutual information with every factor. Thus, $1 - \delta_i$ is used as a modularity score for vector dimension i and the mean of $1 - \delta_i$ over $i$ as the modularity score for the overall representation. Higher values imply better disentanglement.

    \item \textit{Explicitness}~\cite{ridgeway2018learning}: Mean of the ROC area-under-the-curve ($AUC_{jk}$) of a one-versus-rest logistic-regression classifier that takes the latent vectors as input and has factor values as targets, over a factor index $j$ and an index $k$ on values of factor $j$. Represents the coverage of the representation, in other words, how well each factor is represented. Higher values imply better disentanglement.

    \item \textit{Disentanglement Score}~\cite{eastwood2018framework}: The degree to which a representation factorises or disentangles the underlying factors of variation, with each variable (or dimension) capturing at most one generative factor. It is computed as a weighted average of a disentanglement score $D_i = (1 -  H_K(P_i.))$ for each latent dimension variable $c_i$, on the relevance of each $c_i$, where $H_K(P_i.)$ denotes the entropy and $P_{ij}$ denotes the 'probability' of $c_i$ being important for predicting $z_j$. If $c_i$ is important for predicting a single generative factor, the score will be 1. If $c_i$ is equally important for predicting all generative factors, the score will be 0. Higher values imply better disentanglement.

    \item \textit{Completeness Score}~\cite{eastwood2018framework}: The degree to which each underlying factor is captured by a single latent dimension variable. For a given $z_j$ it is given by $C_j = (1 - H_D(\tilde{P}.j))$, where $H_D(\tilde{P}.j) = -\sum_{d=0}^{D - 1}{\tilde{P}_{dj} log_D \tilde{P}_{ij}}$ denotes the entropy of the $\tilde{P}.j$ distribution. If a single latent dimension variable contributes to $z_j$'s prediction, the score will be 1 (complete). If all code variables contribute equally to $z_j$'s prediction, the score will be 0 (maximally over-complete). Higher values imply better disentanglement.

    \item \textit{Informativeness Score}~\cite{eastwood2018framework}: The amount of information that a representation captures about the underlying factors of variation. Given a latent representation $c$, It is quantified for each generative factor $z_j$ by the prediction error $E(z_j , \hat{z}_j)$ (averaged over the dataset), where $E$ is an appropriate error function and $\hat{z}_j = f_j(c)$. Lower values imply better disentanglement.
\end{enumerate}

%% file: tables/optimus_def_examples.tex
\begin{table}[h]
    \small \setlength\tabcolsep{4.5pt}
    \centering
    \begin{tabular}{|p{1.5cm}|p{6cm}|}  \hline
        \textbf{Word}         & \textbf{Generated Definition}                       \\ \hline
        Fox & a member of the Mayflower \\ \hline
        Untermeyer & United States writer of short stories \\ \hline
        organise & make logical or comprehensible \\ \hline
        dishrag & remove the fur from \\ \hline
        altocumulus cloud & a clear blue sky \\ \hline
        shuffle & move quickly on or move quickly forward \\ \hline
        sharpen & make sharp or sharper \\ \hline
        semantic error & discrimination that invalidates an earlier characteristic \\ \hline
        railway station & station where planes take off and land or take off \\ \hline
        Antonio Pignatelli & Italian cardinal and theologian \\ \hline
        union & a cooperative level of play in league with other players \\ \hline
        love knot & a knot of contrasting color or yarn used for tying a wedding band \\ \hline
        commodity brokerage & a place where stockbrokers sell their stock \\ \hline
    \end{tabular}
    \caption{Generation definitions from the Optimus-based model.} \label{tab:srl_silva}
\end{table}

%% file: eacl2023.bbl
\begin{thebibliography}{37}
\expandafter\ifx\csname natexlab\endcsname\relax\def\natexlab#1{#1}\fi

\bibitem[{Bahdanau et~al.(2017)Bahdanau, Bosc, Jastrz{\k{e}}bski, Grefenstette,
  Vincent, and Bengio}]{bahdanau2017learning}
Dzmitry Bahdanau, Tom Bosc, Stanis{\l}aw Jastrz{\k{e}}bski, Edward
  Grefenstette, Pascal Vincent, and Yoshua Bengio. 2017.
\newblock Learning to compute word embeddings on the fly.
\newblock \emph{arXiv preprint arXiv:1706.00286}.

\bibitem[{Bao et~al.(2019)Bao, Zhou, Huang, Li, Mou, Vechtomova, Dai, and
  Chen}]{bao2019generating}
Yu~Bao, Hao Zhou, Shujian Huang, Lei Li, Lili Mou, Olga Vechtomova, Xinyu Dai,
  and Jiajun Chen. 2019.
\newblock Generating sentences from disentangled syntactic and semantic spaces.
\newblock In \emph{Proceedings of the 57th Annual Meeting of the Association
  for Computational Linguistics}, pages 6008--6019.

\bibitem[{Bengio et~al.(2013)Bengio, Courville, and
  Vincent}]{Bengio2013RepresentationLA}
Yoshua Bengio, Aaron~C. Courville, and Pascal Vincent. 2013.
\newblock Representation learning: A review and new perspectives.
\newblock \emph{IEEE Transactions on Pattern Analysis and Machine
  Intelligence}, 35:1798--1828.

\bibitem[{Bevilacqua et~al.(2020)Bevilacqua, Maru, and
  Navigli}]{bevilacqua2020generationary}
Michele Bevilacqua, Marco Maru, and Roberto Navigli. 2020.
\newblock Generationary or:“how we went beyond word sense inventories and
  learned to gloss”.
\newblock In \emph{Proceedings of the 2020 Conference on Empirical Methods in
  Natural Language Processing (EMNLP)}, pages 7207--7221.

\bibitem[{Bosc and Vincent(2018)}]{bosc2018auto}
Tom Bosc and Pascal Vincent. 2018.
\newblock Auto-encoding dictionary definitions into consistent word embeddings.
\newblock In \emph{Proceedings of the 2018 Conference on Empirical Methods in
  Natural Language Processing}, pages 1522--1532.

\bibitem[{Bowman et~al.(2016)Bowman, Vilnis, Vinyals, Dai, Jozefowicz, and
  Bengio}]{bowman2016generating}
Samuel Bowman, Luke Vilnis, Oriol Vinyals, Andrew Dai, Rafal Jozefowicz, and
  Samy Bengio. 2016.
\newblock Generating sentences from a continuous space.
\newblock In \emph{Proceedings of The 20th SIGNLL Conference on Computational
  Natural Language Learning}, pages 10--21.

\bibitem[{Carbonneau et~al.(2022)Carbonneau, Zaidi, Boilard, and
  Gagnon}]{carbonneau2022measuring}
Marc-Andr{\'e} Carbonneau, Julian Zaidi, Jonathan Boilard, and Ghyslain Gagnon.
  2022.
\newblock Measuring disentanglement: A review of metrics.
\newblock \emph{IEEE Transactions on Neural Networks and Learning Systems}.

\bibitem[{Chen et~al.(2019)Chen, Tang, Wiseman, and Gimpel}]{chen2019multi}
Mingda Chen, Qingming Tang, Sam Wiseman, and Kevin Gimpel. 2019.
\newblock A multi-task approach for disentangling syntax and semantics in
  sentence representations.
\newblock In \emph{NAACL}.

\bibitem[{Chen et~al.(2018)Chen, Li, Grosse, and Duvenaud}]{chen2018isolating}
Ricky~TQ Chen, Xuechen Li, Roger Grosse, and David Duvenaud. 2018.
\newblock Isolating sources of disentanglement in vaes.
\newblock In \emph{Proceedings of the 32nd International Conference on Neural
  Information Processing Systems}, pages 2615--2625.

\bibitem[{Cheng et~al.(2020)Cheng, Min, Shen, Malon, Zhang, Li, and
  Carin}]{cheng2020improving}
Pengyu Cheng, Martin~Renqiang Min, Dinghan Shen, Christopher Malon, Yizhe
  Zhang, Yitong Li, and Lawrence Carin. 2020.
\newblock Improving disentangled text representation learning with
  information-theoretic guidance.
\newblock In \emph{Proceedings of the 58th Annual Meeting of the Association
  for Computational Linguistics}, pages 7530--7541.

\bibitem[{Colombo et~al.(2021)Colombo, Piantanida, and
  Clavel}]{colombo2021novel}
Pierre Colombo, Pablo Piantanida, and Chlo{\'e} Clavel. 2021.
\newblock A novel estimator of mutual information for learning to disentangle
  textual representations.
\newblock In \emph{Proceedings of the 59th Annual Meeting of the Association
  for Computational Linguistics}, pages 6539--6550.

\bibitem[{Dupont(2018)}]{dupont2018learning}
Emilien Dupont. 2018.
\newblock Learning disentangled joint continuous and discrete representations.
\newblock \emph{Advances in Neural Information Processing Systems}, 31.

\bibitem[{Eastwood and Williams(2018)}]{eastwood2018framework}
Cian Eastwood and Christopher~KI Williams. 2018.
\newblock A framework for the quantitative evaluation of disentangled
  representations.
\newblock In \emph{6th International Conference on Learning Representations}.

\bibitem[{Gadetsky et~al.(2018)Gadetsky, Yakubovskiy, and
  Vetrov}]{gadetsky2018conditional}
Artyom Gadetsky, Ilya Yakubovskiy, and Dmitry Vetrov. 2018.
\newblock Conditional generators of words definitions.
\newblock In \emph{Proceedings of the 56th Annual Meeting of the Association
  for Computational Linguistics (Volume 2: Short Papers)}, pages 266--271.

\bibitem[{Higgins et~al.(2017)Higgins, Matthey, Pal, Burgess, Glorot,
  Botvinick, Mohamed, and Lerchner}]{higgins2016beta}
Irina Higgins, Lo{\"i}c Matthey, Arka Pal, Christopher~P. Burgess, Xavier
  Glorot, Matthew~M. Botvinick, Shakir Mohamed, and Alexander Lerchner. 2017.
\newblock beta-vae: Learning basic visual concepts with a constrained
  variational framework.
\newblock In \emph{ICLR}.

\bibitem[{Hill et~al.(2016)Hill, Cho, Korhonen, and Bengio}]{hill2016learning}
Felix Hill, KyungHyun Cho, Anna Korhonen, and Yoshua Bengio. 2016.
\newblock Learning to understand phrases by embedding the dictionary.
\newblock \emph{Transactions of the Association for Computational Linguistics},
  4:17--30.

\bibitem[{Hu et~al.(2017)Hu, Yang, Liang, Salakhutdinov, and
  Xing}]{hu2017toward}
Zhiting Hu, Zichao Yang, Xiaodan Liang, Ruslan Salakhutdinov, and Eric~P Xing.
  2017.
\newblock Toward controlled generation of text.
\newblock In \emph{International Conference on Machine Learning}, pages
  1587--1596. PMLR.

\bibitem[{John et~al.(2019)John, Mou, Bahuleyan, and
  Vechtomova}]{john2019disentangled}
Vineet John, Lili Mou, Hareesh Bahuleyan, and Olga Vechtomova. 2019.
\newblock Disentangled representation learning for non-parallel text style
  transfer.
\newblock In \emph{Proceedings of the 57th Annual Meeting of the Association
  for Computational Linguistics}, pages 424--434.

\bibitem[{Kim and Mnih(2018)}]{kim2018disentangling}
Hyunjik Kim and Andriy Mnih. 2018.
\newblock \href {https://proceedings.mlr.press/v80/kim18b.html} {Disentangling
  by factorising}.
\newblock In \emph{Proceedings of the 35th International Conference on Machine
  Learning}, volume~80 of \emph{Proceedings of Machine Learning Research},
  pages 2649--2658. PMLR.

\bibitem[{Kingma and Welling(2014)}]{Kingma2014AutoEncodingVB}
Diederik~P. Kingma and Max Welling. 2014.
\newblock Auto-encoding variational bayes.

\bibitem[{Li et~al.(2020)Li, Gao, Li, Peng, Li, Zhang, and Gao}]{li2020optimus}
Chunyuan Li, Xiang Gao, Yuan Li, Baolin Peng, Xiujun Li, Yizhe Zhang, and
  Jianfeng Gao. 2020.
\newblock Optimus: Organizing sentences via pre-trained modeling of a latent
  space.
\newblock In \emph{Proceedings of the 2020 Conference on Empirical Methods in
  Natural Language Processing (EMNLP)}, pages 4678--4699.

\bibitem[{Locatello et~al.(2019)Locatello, Bauer, Lucic, Raetsch, Gelly,
  Sch{\"o}lkopf, and Bachem}]{locatello2019challenging}
Francesco Locatello, Stefan Bauer, Mario Lucic, Gunnar Raetsch, Sylvain Gelly,
  Bernhard Sch{\"o}lkopf, and Olivier Bachem. 2019.
\newblock Challenging common assumptions in the unsupervised learning of
  disentangled representations.
\newblock In \emph{international conference on machine learning}, pages
  4114--4124. PMLR.

\bibitem[{McInnes et~al.(2018)McInnes, Healy, and Melville}]{mcinnes2018umap}
Leland McInnes, John Healy, and James Melville. 2018.
\newblock Umap: Uniform manifold approximation and projection for dimension
  reduction.
\newblock \emph{arXiv preprint arXiv:1802.03426}.

\bibitem[{Mercatali and Freitas(2021)}]{mercatali2021disentangling}
Giangiacomo Mercatali and Andr{\'e} Freitas. 2021.
\newblock Disentangling generative factors in natural language with discrete
  variational autoencoders.
\newblock In \emph{Findings of the Association for Computational Linguistics:
  EMNLP 2021}, pages 3547--3556.

\bibitem[{Merity et~al.(2016)Merity, Xiong, Bradbury, and
  Socher}]{merity2016pointer}
Stephen Merity, Caiming Xiong, James Bradbury, and Richard Socher. 2016.
\newblock Pointer sentinel mixture models.
\newblock \emph{arXiv preprint arXiv:1609.07843}.

\bibitem[{Noraset et~al.(2017)Noraset, Liang, Birnbaum, and
  Downey}]{noraset2017definition}
Thanapon Noraset, Chen Liang, Larry Birnbaum, and Doug Downey. 2017.
\newblock Definition modeling: Learning to define word embeddings in natural
  language.
\newblock In \emph{Thirty-First AAAI Conference on Artificial Intelligence}.

\bibitem[{Papineni et~al.(2002)Papineni, Roukos, Ward, and
  Zhu}]{papineni2002bleu}
Kishore Papineni, Salim Roukos, Todd Ward, and Wei-Jing Zhu. 2002.
\newblock Bleu: a method for automatic evaluation of machine translation.
\newblock In \emph{Proceedings of the 40th annual meeting of the Association
  for Computational Linguistics}, pages 311--318.

\bibitem[{Raffel et~al.(2020)Raffel, Shazeer, Roberts, Lee, Narang, Matena,
  Zhou, Li, and Liu}]{Raffel2020t5}
Colin Raffel, Noam Shazeer, Adam Roberts, Katherine Lee, Sharan Narang, Michael
  Matena, Yanqi Zhou, Wei Li, and Peter~J. Liu. 2020.
\newblock \href {http://jmlr.org/papers/v21/20-074.html} {Exploring the limits
  of transfer learning with a unified text-to-text transformer}.
\newblock \emph{Journal of Machine Learning Research}, 21(140):1--67.

\bibitem[{Ridgeway and Mozer(2018)}]{ridgeway2018learning}
Karl Ridgeway and Michael~C Mozer. 2018.
\newblock Learning deep disentangled embeddings with the f-statistic loss.
\newblock In \emph{Proceedings of the 32nd International Conference on Neural
  Information Processing Systems}, pages 185--194.

\bibitem[{Shen et~al.(2020)Shen, Mueller, Barzilay, and
  Jaakkola}]{shen2020educating}
Tianxiao Shen, Jonas Mueller, Regina Barzilay, and Tommi Jaakkola. 2020.
\newblock Educating text autoencoders: Latent representation guidance via
  denoising.
\newblock In \emph{International Conference on Machine Learning}, pages
  8719--8729. PMLR.

\bibitem[{Silva et~al.(2016)Silva, Handschuh, and
  Freitas}]{silva2016categorization}
Vivian Silva, Siegfried Handschuh, and Andr{\'e} Freitas. 2016.
\newblock Categorization of semantic roles for dictionary definitions.
\newblock In \emph{Proceedings of the 5th Workshop on Cognitive Aspects of the
  Lexicon (CogALex-V)}, pages 176--184.

\bibitem[{Silva et~al.(2018)Silva, Handschuh, and
  Freitas}]{silva2018recognizing}
Vivian~S Silva, Siegfried Handschuh, and Andr{\'e} Freitas. 2018.
\newblock Recognizing and justifying text entailment through distributional
  navigation on definition graphs.
\newblock In \emph{Thirty-Second AAAI Conference on Artificial Intelligence}.

\bibitem[{Tsukagoshi et~al.(2021)Tsukagoshi, Sasano, and
  Takeda}]{Tsukagoshi2021DefSentSE}
Hayato Tsukagoshi, Ryohei Sasano, and Koichi Takeda. 2021.
\newblock Defsent: Sentence embeddings using definition sentences.
\newblock In \emph{ACL/IJCNLP}.

\bibitem[{Van~der Maaten and Hinton(2008)}]{van2008visualizing}
Laurens Van~der Maaten and Geoffrey Hinton. 2008.
\newblock Visualizing data using t-sne.
\newblock \emph{Journal of machine learning research}, 9(11).

\bibitem[{Vishnubhotla et~al.(2021)Vishnubhotla, Hirst, and
  Rudzicz}]{vishnubhotla2021evaluation}
Krishnapriya Vishnubhotla, Graeme Hirst, and Frank Rudzicz. 2021.
\newblock An evaluation of disentangled representation learning for texts.
\newblock In \emph{Findings of the Association for Computational Linguistics:
  ACL-IJCNLP 2021}, pages 1939--1951.

\bibitem[{Zhang et~al.(2021)Zhang, Prokhorov, and
  Shareghi}]{zhang2021unsupervised}
Lan Zhang, Victor Prokhorov, and Ehsan Shareghi. 2021.
\newblock Unsupervised representation disentanglement of text: An evaluation on
  synthetic datasets.
\newblock In \emph{Proceedings of the 6th Workshop on Representation Learning
  for NLP (RepL4NLP-2021)}.

\bibitem[{Zhao et~al.(2017)Zhao, Zhao, and Eskenazi}]{zhao2017learning}
Tiancheng Zhao, Ran Zhao, and Maxine Eskenazi. 2017.
\newblock Learning discourse-level diversity for neural dialog models using
  conditional variational autoencoders.
\newblock In \emph{Proceedings of the 55th Annual Meeting of the Association
  for Computational Linguistics (Volume 1: Long Papers)}, pages 654--664.

\end{thebibliography}
